\RequirePackage{fix-cm}
\documentclass[smallextended]{svjour3}       
\smartqed  
\usepackage{graphicx}
\usepackage[colorlinks,linkcolor=red]{hyperref}   
\usepackage{bigstrut,multirow,rotating}      
\usepackage{booktabs}                             
\usepackage{enumerate}
\usepackage{amsmath}
\usepackage{bbm}
\usepackage{amsfonts,amssymb}
\usepackage{appendix}
\usepackage[numbers,sort&compress]{natbib}
\usepackage{url}
\usepackage[hyphenbreaks]{breakurl}

\begin{document}

\title{Extracting stochastic dynamical systems with $\alpha$-stable L\'evy noise from data
}

\titlerunning{Extracting stochastic dynamical systems from data}        

\author{Yang Li        \and
        Yubin Lu       \and
        Shengyuan Xu   \and
        Jinqiao Duan
}


\institute{Li Y. and Xu S. \at
               School of Automation, Nanjing University of Science and Technology, Nanjing 210094, China           
           \and
           Lu Y. \at
             School of Mathematics and Statistics \& Center for Mathematical Sciences, Huazhong University of Science and Technology, Wuhan 430074, China
           \and
           Duan J. \at
             Departments of Applied Mathematics \& Physics, Illinois Institute of Technology, Chicago, IL 60616, USA\\
             \\
\email{liyangbx5433@163.com}
}

\date{Received: date / Accepted: date}

\maketitle

\begin{abstract}
With the rapid increase of valuable observational, experimental and simulated data for complex systems, much efforts have been devoted to identifying governing laws underlying the evolution of these systems. Despite the wide applications of non-Gaussian fluctuations in numerous physical phenomena, the data-driven approaches to extract stochastic dynamical systems with (non-Gaussian) L\'evy noise are relatively few so far. In this work, we propose a data-driven method to extract stochastic dynamical systems with $\alpha$-stable L\'evy noise from short burst data based on the properties of $\alpha$-stable distributions. More specifically, we first estimate the L\'evy jump measure and noise intensity via computing mean and variance of the amplitude of the increment of the sample paths. Then we approximate the drift coefficient by combining nonlocal Kramers-Moyal formulas with normalizing flows. Numerical experiments on one- and two-dimensional prototypical examples illustrate the accuracy and effectiveness of our method. This approach will become an effective scientific tool in discovering stochastic governing laws of complex phenomena and understanding dynamical behaviors under non-Gaussian fluctuations.

\end{abstract}

\section{Introduction}
\label{intro}
Newton's second law or other physical laws are essential to establish the mathematical model for studying a new physical phenomenon. From this point of view, dynamical modeling requires a deep understanding of the process to be analyzed. The essence of model abstraction is an approximation to the observed reality, which is usually represented by a system composed of ordinary or partial differential equations, deterministic or stochastic differential equations, and control equations. Although mathematical models are accurate for many processes, it is particularly difficult to develop such models for some of the most challenging systems, including climate dynamics, brain dynamics, biological systems and financial markets.

Fortunately, more and more data are observed or measured in recent years with the development of scientific tools and simulation capabilities. Therefore, a large number of data-driven methods has been proposed to discover governing laws of systems from data. For instance, several researchers designed the Sparse Identification of Nonlinear Dynamics approach to extract deterministic ordinary \cite{SINDy1} or partial \cite{Kevrekidis1, Schaeffer2013, RudyKutz2019} differential equations from available path data. Subsequently, Boninsegna et al. \cite{Boninsegna2018} extended this method to learn It\^o stochastic differential equations via Kramers–Moyal expansions. Based on the theory of Koopman operator, Klus et al. \cite{SKO3} generalized the Extended Dynamic Mode Decomposition method \cite{EDMD} to identify stochastic differential equations from sample path data. There are also other data-driven methods such as neural differential equations \cite{Duvenaud1, Duvenaud2, NeuralSDE, Felix} and Bayesian inference \cite{GPs1,GPs2} to extract stochastic differential equations. These methods are only applicable to extract either deterministic differential equations or stochastic differential equations with Gaussian noise.

However, there are many complex phenomena involving jump, bursting, intermittent and hopping, which are more appropriate to be modeled as stochastic dynamical systems with non-Gaussian fluctuations. Theoretically, a significant class of Markov processes (so called Feller processes) can be modeled by stochastic differential equations with Brownian motion and L\'evy motion together \cite{Bottcher}. According to the Greenland ice core measurement data, Ditlevsen \cite{Ditlevsen} found that the temperature evolution in the climate system can be described via a stochastic differential equation with $\alpha $-stable L\'evy motion. Subsequently, Zheng et al. \cite{ZhengYY2020} developed a probabilistic framework to investigate the maximum likelihood climate change for an energy balance system under the combined influence of greenhouse effect and $\alpha$-stable Lévy motions. Additionally, there are also some researchers using the L\'evy motion to characterize the random fluctuations emerged in neural systems \cite{NeuralModel}, gene networks \cite{GeneModel}, epidemic model \cite{LIU201829}, and the Earth systems \cite{ashwin2012tipping, YangDuanWiggins2020, WeiPY2020, DannyTesfay, liyang2020b}.

Therefore, stochastic dynamical systems with L\'evy motion are mathematical models with profound scientific significance. It is necessary and desirable to develop data-driven approaches to extract non-Gaussian stochastic governing laws. So far, there are a few methods proposed recently. The first method is by nonlocal Kramers-Moyal formulas to express the L\'evy jump measure, drift coefficient and diffusion coefficient by transition probability density (solution of nonlocal Fokker-Planck equation) or sample paths (solution of stochastic differential equation) \cite{YangLi2020a, liyang2021b}, which can be also realized via normalizing flows \cite{luy2021a}. The second method is to learn the nonlocal Fokker-Planck equation corresponding to the stochastic differential equation by neural networks \cite{XiaoliChen}. In the third method, the coefficients of the stochastic differential equations can be estimated by generalizing the method of Koopman operator to non-Gaussian case \cite{LuYB2020}.

In this present work, we devise a new data-driven method to learn stochastic dynamical systems with $\alpha$-stable L\'evy noise from short burst data, based on the properties of $\alpha$-stable distribution. The L\'evy stability parameter and noise intensity can be estimated via the mean and variance of the amplitude of the increments of sample paths. The drift coefficient can be identified by nonlocal Kramers-Moyal formulas combined with normalizing flows. The idea of normalizing flows was introduced by Tabak and Vanden-Eijnden \cite{tabak-2010}, which is devoted to transforming a given distribution into a standard normal one via an inverse map. After choosing a suitable transformation, the estimate of target probability density can be obtained.

This article is arranged as follows. In Section \ref{prel}, we review the basic definitions and properties of $\alpha$-stable distribution and L\'evy motion. In Section \ref{theo}, we prove a theorem to express the relations between the L\'evy jump measure, noise intensity and drift coefficient with the sample paths of stochastic dynamical systems. Then we design numerical algorithms to extract the stochastic dynamical systems based on this theorem in Section \ref{nume}. Several prototypical examples are tested to illustrate the effectiveness of our method in Section \ref{exam}. Finally, the discussions are presented in Section \ref{disc}.

\section{Preliminaries}
\label{prel}

\subsection{The $\alpha$-stable distribution}
\label{stable}
The $\alpha$-stable distributions are a rich class of probability distributions that allow skewness and heavy tails and have many intriguing mathematical properties. According to definition \cite{Duan2015}, a random variable $X$ is called a stable random variable if it is a limit in distribution of a scaled sequence $\left( S_n-b_n \right) / a_n$, where $S_n=X_1+\cdots+X_n$, $X_i$ are some independent identically distributed random variables and $a_n>0$ and $b_n$ are some real sequences.

A univariate stable random variable $X$ with stability parameter $\alpha$, skewness parameter $\beta$, scaling parameter $\gamma$, shift parameter $\delta$ has characteristic function
\begin{align}\label{stableCF}
\mathbb{E} \text{exp} \left( iuX \right) = \text{exp} \left\{ -{\gamma}^{\alpha} \left[ {\left| u \right|}^{\alpha} + i\beta \eta \left( u,\alpha \right) \right] + iu\delta \right\},
\end{align}
where $0<\alpha<2$, $-1 \le \beta \le 1$, $\gamma>0$, $\delta \in \mathbb{R}$ and
$$
\eta \left( u,\alpha \right) = \left\{ \begin{array}{ccc}
   -\left( \text{sign} (u) \text{tan} \left( \pi \alpha /2 \right) \right) {\left| u \right|}^{\alpha}, & \alpha \ne 1,  \\
   \left( 2/\pi \right) u \ \text{ln} \left| u \right|, & \alpha =1.  \\
\end{array} \right.
$$
The distribution of a stable random variable is denoted as ${S}_{\alpha} \left( \gamma, \beta, \delta \right)$. When $\alpha=2$ and $\beta=0$, the stable random variable corresponds to Gaussian random variable.

Let $\mathbb{S}$ be the unit sphere in $\mathbb{R}^{n}: \mathbb{S} = \left \{ v \in \mathbb{R}^{n}: \left|v \right| =1 \right \}$. Then a stable random vector can be characterized by a spectral measure ${\rm \Lambda}$ (a finite measure on $\mathbb{S}$) and a shift vector $\delta \in \mathbb{R}^{n}$ \cite{Applebaum, nolan2005}. We will write $X \in {S}_{\alpha} \left( {\rm \Lambda}, \delta \right)$ if the joint characteristic function of $X$ is
\begin{align}\label{stableVCF}
\mathbb{E} \text{exp} \left( i\langle u, X \rangle \right) = \text{exp} \left\{ -\int_{\mathbb{S}} {\left[ {\left| \langle u, s \rangle \right|}^{\alpha} + i\eta \left( \langle u, s \rangle, \alpha \right) \right] {\rm \Lambda} \left( ds \right)} + i\langle u, \delta \rangle \right\}.
\end{align}
where $0<\alpha<2$.

If the spectral measure is a uniform distribution on $\mathbb{S}$, then $X$ is called rotationally symmetric or isotropic with characteristic function
\begin{align}\label{stableRSCF}
\mathbb{E} \text{exp} \left( i\langle u, X \rangle \right)  = \text{exp} \left\{ -{\gamma}^{\alpha} {\left| u \right|}^{\alpha}  \right\}.
\end{align}
According to Nolan \cite{Nolan}, the rotationally symmetric $\alpha$-stable distributions are scale mixtures of multivariate normal distributions. Specifically, let $A \sim {S}_{\alpha /2} \left( 2{\gamma}^{2} {\left( \text{cos} \left( \pi \alpha/4 \right) \right)} ^ {2/{\alpha}}, 1, 0 \right)$ be a stable random variable, and $G \sim N \left( 0, I \right)$ be an $n$-dimensional standard normal random vector which is independent of $A$. Then the random vector $X=A^{1/2}G$ is rotationally symmetric with characteristic function (\ref{stableRSCF}).

Let $X$ be an $n$-dimensional rotationally symmetric stable random vector with characteristic function (\ref{stableRSCF}). Then the amplitude of $X$ is defined as
\begin{align}\label{ampli}
R= \left| X \right|= \sqrt{{X}_{1}^{2}+ \cdots +{X}_{n}^{2}}.
\end{align}
Then we have 
\begin{align}\label{ampli2}
R^2 \overset{d}{=} A \left( {G}_{1}^{2}+ \cdots +{G}_{n}^{2} \right),
\end{align}
where the symbol $\overset{d}{=}$ denotes the same distribution. Consequently, the fractional moments of $R$ can be derived using (\ref{ampli2}): if $-n<p<\alpha$,
\begin{equation}\label{framo}
\begin{split}
\mathbb{E} \left[ R^p \right] &= \mathbb{E} \left[ A^{p/2} {\left( {G}_{1}^{2}+ \cdots +{G}_{n}^{2} \right)}^{p/2} \right] \\ &= \mathbb{E} \left[ A^{p/2}\right] \mathbb{E} \left[ {\left( {G}_{1}^{2}+ \cdots +{G}_{n}^{2} \right)}^{p/2} \right]  \\ &= {\left( 2\gamma \right)}^{p} \frac{\Gamma \left( 1-p/\alpha \right)}{\Gamma \left( 1-p/2 \right)} \frac{\Gamma \left( \left( n+p \right)/2 \right)}{\Gamma \left( n/2 \right)},
\end{split}
\end{equation}
where the Gamma function is defined by $\Gamma \left( z \right)= \int_{0}^{\infty}{{t}^{z-1} e^{-t} dt}$.

According to the formula (\ref{framo}), we can derive the following lemma \cite{Nolan}:
\newtheorem{thm}{\bf Theorem}
\newtheorem{lem}[thm]{\bf Lemma}
\begin{lem} \label{lem1}
$ {\rm ln} R $ has moment generating function $\mathbb{E} {\rm exp} \left( u {\rm ln} R \right)= \mathbb{E} R^u$ given by (\ref{framo}). The mean and variance of ${\rm ln}R$ are
\begin{align}\label{L1}
\mathbb{E} \left( {\rm ln}R \right)= {\rm ln} \left( 2\gamma \right) + {\gamma}_{\rm Euler} \left( \frac{1}{\alpha} - \frac{1}{2} \right) + \frac{1}{2} \psi \left( n/2 \right),
\end{align}
\begin{align}\label{L2}
{\rm Var} \left( {\rm ln}R \right)= \frac{{\pi}^2}{6} \left( \frac{1}{{\alpha}^2} - \frac{1}{4} \right) + \frac{1}{4} {\psi}^{'} \left( n/2 \right),
\end{align}
where the digamma function $\psi \left( z \right)= {\Gamma}^{'} \left( z \right) / {\Gamma} \left( z \right)$ and the Euler's constant ${\gamma}_{\rm Euler}= -{\Gamma}^{'} \left( 1 \right) \approx 0.57721$.
\end{lem}

\subsection{L\'evy motion}
\label{Levy}
Let $L_{t}$ be a stochastic process in $\mathbb{R}^n$ defined on a probability space $(\Omega, \mathcal{F}, P)$. We say that $L_{t}$ is a L\'evy motion if:\\
(1) $L_{0}=0$, a.s.;\\
(2) $L_{t}$ has independent and stationary increments;\\
(3) $L_{t}$ is stochastically continuous, i.e., for all $\delta>0$ and for all $s\geq0$
$$
\lim\limits_{t\to s}P(|L_{t}-L_{s}|>\delta)=0.
$$

For a L\'evy motion $(L_{t},t\geq0)$, we have the L\'evy-Khinchine formula
\begin{align*}
& \mathbb{E} \text{exp} \left[ i\langle u,L_t \rangle \right] = \text{exp} \left[ t \eta \left( u \right) \right],\\
& \eta \left( u \right) = i\langle b,u \rangle -\frac{1}{2} \langle u,Au \rangle  + \int_{\mathbb{R}^n \backslash\{0\}} \left[ e^{i \langle u,y \rangle}-1-i \langle u,y \rangle I_{ \left\{\Vert y\Vert<1 \right\}}(y) \right]\, \nu(dy)
\end{align*}
for each $t\geq0$, $u\in\mathbb{R}^n$, where $(b,A,\nu)$ is the triplet of L\'evy motion $L_{t}$. Usually, we consider the pure jump case $\left( 0,0,\nu \right)$.

The rotationally symmetric $\alpha$-stable L\'evy motion is a special but popular type of the L\'evy process which is defined by the rotationally symmetric stable random vector. Its jump measure has the form
\begin{align}
\label{jumpm}
\nu \left( dy \right) = c \left( n, \alpha \right) {\left| y \right|} ^ {-n-\alpha} dy
\end{align}
for $y \in \mathbb{R}^{n} \backslash \{0\}$ with the intensity constant
\begin{align}
\label{jumpc}
c \left( n, \alpha \right) = \frac {\alpha \Gamma \left( \left( n+\alpha \right)/2 \right)} {2^{1-\alpha} {\pi}^{n/2} \Gamma \left( 1-\alpha/2 \right)}
\end{align}
It has larger jumps with lower jump frequencies for smaller $\alpha$ ($0<\alpha <1$), while it has smaller jump sizes with higher jump frequencies for larger $\alpha$ ($1<\alpha <2$). The special case $\alpha=2$ corresponds to (Gaussian) Brownian motion. For more information about L\'evy motion, refer to References \cite{Duan2015, Applebaum}.

\section{Theory and method}
\label{theo}
Consider an $n$-dimensional stochastic dynamical system
\begin{equation}
\label{sde}
dx\left( t \right)=b\left(x\left( t \right) \right)dt+\sigma d{{L}_{t}},
\end{equation}
where ${{L}_{t}}={{\left( {{L}_{1,t}},\ \cdots ,\ {{L}_{n,t}} \right)}^{\text{T}}}$ is an $n$-dimensional rotationally symmetric L\'evy process with the jump measure (\ref{jumpm}) described in the previous section. In this work, we only consider the small jump case $1<\alpha<2$. The vector $b\left(x \right) = {{\left[ {{b}_{1}}\left( x \right),\ \cdots ,\ {{b}_{n}}\left( x \right) \right]}^{\text{T}}}$  is the drift coefficient (or vector field) in ${{\mathbb{R}}^{n}}$.  We take the positive constant $\sigma $  as  the noise intensity of the L\'evy process and assume that the initial condition is $x\left( 0 \right) =z$.

This research is devoted to proposing a data-driven method to extract the stochastic dynamical system as the form (\ref{sde}) from sample paths. Essentially, we need to identify the L\'evy jump measure or the stability parameter $\alpha$, the L\'evy noise intensity $\sigma$, and the drift coefficient $b \left( x \right)$. 

Based on our previous work, it is shown that the drift coefficient can be expressed by the transition probability density function (solution of nonlocal Fokker-Planck equation) based on nonlocal Kramers-Moyal formulas \cite{YangLi2020a}. Its key idea of computing the drift term is to limit the integration domain inside a spherical surface instead of the whole space in the usual Kramers-Moyal formulas of Gaussian case to avoid its divergence successfully. Next we present how to compute the L\'evy jump measure and noise intensity in the following, according to the properties of rotationally symmetric stable random vector in Section \ref{stable}.

For small $t>0$, the equation (\ref{sde}) can be approximately rewritten as
\begin{equation}
\label{sde2}
x\left( t \right)-z  \approx b\left(z \right)t+\sigma {{L}_{t}}.
\end{equation}
Since the characteristic function of this L\'evy motion $L_{t}$ has the form
\begin{equation}
\label{rsLCF}
\mathbb{E} \text{exp} \left( i\langle u, {L}_{t} \rangle \right)= \text{exp} \left( -t {\left| u \right|}^{\alpha} \right),
\end{equation}
we have
\begin{equation}
\label{LtL1}
t^{-1/\alpha} L_{t} \overset{d}{=} L_{1}.
\end{equation}
Substituting it to Eq. (\ref{sde2}) yields
\begin{equation}
\label{sde3}
{\sigma}^{-1} t^{-1/\alpha} \left( x\left( t \right) - z \right)  \approx {\sigma}^{-1} b\left( z \right) {t}^{1-1/\alpha}+ {{L}_{1}}.
\end{equation}
When $\alpha \in \left( 1, 2 \right)$, we can obtain
\begin{equation}
\label{sde4}
\underset{t\to 0}{\mathop{\lim }} {\sigma}^{-1} t^{-1/\alpha} \left( x\left( t \right) - z \right) \overset{d}{=} {{L}_{1}}.
\end{equation}
Since $L_{1}$ is a standard rotationally symmetric $\alpha$-stable random vector, the variable ${\rm ln} \left| L_{1} \right|$ satisfies Lemma \ref{lem1}.

In conclusion, we have the following theorem:

\newtheorem{thm2}[thm]{\bf Theorem}
\begin{thm2} \label{thm2}
The L\'evy jump measure, the L\'evy noise intensity and the drift coefficient have the following relations with the sample paths of stochastic dynamical equation (\ref{sde}):\\
1) 
\begin{align}\label{T1}
\begin{split}
& \underset{t\to 0}{\mathop{\lim }} \ \mathbb{E}  \left( {\rm ln} \left| x\left( t \right) - z \right| -\frac{1}{\alpha} {\rm ln}t \right) \\& = {\rm ln} \left( 2\sigma \right) + {\gamma}_{\rm Euler} \left( \frac{1}{\alpha} - \frac{1}{2} \right) + \frac{1}{2} \psi \left( n/2 \right),
\end{split}
\end{align}
\begin{align}\label{T2}
\begin{split}
& \underset{t\to 0}{\mathop{\lim }} \ {\rm Var} \left( {\rm ln} \left| x\left( t \right) - z \right| -\frac{1}{\alpha} {\rm ln}t \right) \\&= \frac{{\pi}^2}{6} \left( \frac{1}{{\alpha}^2} - \frac{1}{4} \right) + \frac{1}{4} {\psi}^{'} \left( n/2 \right).
\end{split}
\end{align}
2) For $i=1,\ 2,\ \ldots ,\ n$ and every $\epsilon>0$,
\begin{align}\label{T3}
\begin{split}
& \underset{t\to 0}{\mathop{\lim }}\,{{t}^{-1}}\int_{\left| x-z \right|<\varepsilon}{\left( {{x}_{i}}-{{z}_{i}} \right)p\left( x,t|z,0 \right) \textrm{d}\mathbf{x}} \\ & = \underset{t\to 0}{\mathop{\lim }}\,{{t}^{-1}}\mathbb{P}\left\{ \left. \left| x\left( t \right)-z \right| <\varepsilon  \right| x\left( 0 \right) =z \right\}\cdot \mathbb{E} \left[ \left. \left( {{x}_{i}}\left( t \right) -{{z}_{i}} \right) \right| x\left( 0 \right)=z;\ \left| x\left( t \right)-z \right| <\varepsilon  \right] \\ & = {{b}_{i}} \left( z \right).
\end{split}
\end{align}
\end{thm2}

Remark that the first assertion of Theorem \ref{thm2} is used to compute the L\'evy jump measure and noise intensity. Two equations (\ref{T1}) and (\ref{T2}) can be used to solve two parameters precisely, the stability paramerter $\alpha$ and the noise intensity $\sigma$. The second assertion of Theorem \ref{thm2} is used to compute the drift coefficient, which stems from the nonlocal Kramers-Moyal formulas. The concrete algorithms are presented in next section.

\section{Numerical algorithms}
\label{nume}
Based on Theorem \ref{thm2}, we design numerical algorithms to identify the stability parameter, noise intensity and drift coefficient from short burst data in this section.

\subsection{Algorithm for identification of the L\'evy motion}
\label{algoLevy}
Assume that there exists a pair of data sets $Z$ and $X$, where $X$ is the image of $Z$ after a small time $t^{*}$ according to the stochastic differential equation (\ref{sde}). Then we can compute the following mean and variance
\begin{equation}\label{mean}
m = \mathbb{E} \left[ \left( {\rm ln} \left| x\left( t^{*} \right) - z \right| \right) \right],
\end{equation}
\begin{equation}\label{var}
V = {\rm Var} \left[ \left( {\rm ln} \left| x\left( t^{*} \right) - z \right| \right) \right].
\end{equation}
Since the constant $-\frac{1}{\alpha} {\rm ln}t^{*}$ does not affect the value of variance and the equation (\ref{T2}) is not related with the noise intensity $\sigma$, the stability parameter $\alpha$ can be derived as
\begin{equation}\label{alpha}
\alpha = \left[ \frac{6}{{\pi}^{2}} \left( V- \frac{1}{4} {\psi}^{'} \left( n/2 \right) \right) + \frac{1}{4} \right] ^{-1/2}.
\end{equation}
Substituting the value $\alpha$ to the equation (\ref{T1}) yields the noise intensity
\begin{equation}\label{sigma}
\sigma = \frac{1}{2} {\rm exp} \left[ m- {\gamma}_{\rm Euler} \left( \frac{1}{\alpha} - \frac{1}{2} \right) -\frac{1}{2} \psi \left( n/2 \right) -\frac{1}{\alpha} {\rm ln}t^{*} \right].
\end{equation}

\subsection{Algorithm for identification of the drift term}
\label{algodrift}
In this subsection, we will introduce a machine learning method, so-called normalizing flows, to estimate probability density from sample data. Subsequently, we will state how to identify the drift term of stochastic differential equation using this estimated probability density via the second assertion of Theorem \ref{thm2}.

Normalizing flows is a generative model in deep learning field, which can be used to express probability density using a prior probability density and a series of bijective transformations \cite{NFs2}.

For a random vector $z\in\mathbb{R}^D$, which satisfies $z\thicksim p_{z}(z)$.
The main idea of normalizing flow is to find a transformation $T$ such that:
\begin{align}
z=T(x), \quad \mbox{where} \quad z\thicksim p_{z}(z),
\end{align}
where $p_{z}(z)$ is a prior density. When the transformation $T$ is invertible and both $T$ and $T^{-1}$ are differentiable, the density $p_{x}(x)$ can be expressed by a change of variables:
\begin{align}\label{CoV}
p_{x}(x)=p_{z}(T(x))\mid {\rm det} J_{T}(x)\mid, \quad \mbox{where} \quad z=T(x)
\end{align}
and $J_{T}(x)$ is the Jacobian of $T$, i.e.,
\begin{center} {$J_{T}(x) = \left[ {\begin{array}{*{20}{c}}
 \frac{\partial T_1}{\partial x_1} & \cdots & \frac{\partial T_1}{\partial x_D}\\
 \vdots & \ddots & \vdots\\
 \frac{\partial T_D}{\partial x_1} & \cdots & \frac{\partial T_D}{\partial x_D}
\end{array}} \right].$}
\end{center}

If we have two invertible and differentiable transformations $T_1$ and $T_2$,  the following properties hold:
\begin{align}
(T_{2}\circ T_{1})^{-1} &= T_{1}^{-1}\circ T_{2}^{-1}, \nonumber\\
{\rm det} J_{T_{2}\circ T_{1}(z)}&={\rm det} J_{T_{2}}(T_{1}(z))\cdot {\rm det} J_{T_{1}}(z). \nonumber
\end{align}
Consequently, we can construct more complex transformations $T$ by these properties, i.e., $T=T_{K}\circ T_{K-1}\circ \cdots \circ T_{1}$, where each $T_k$ transforms $z_{k-1}$ into $z_k$, assuming $z_0=x$ and $z_K=z$.

Given a set of samples $\{x_1, x_2, \ldots, x_N \}$ and it comes from an unknown target density $p_{x}(x;\theta)$. We can minimize the negative log-likelihood on data,
\begin{align}\label{loss1}
\mathcal{L} = -\sum_{i=1}^N {\rm log} p_{x}(x_i; \theta).
\end{align}
Changing variables formula (\ref{CoV}) into (\ref{loss1}), we have
\begin{align}\label{loss2}
\mathcal{L} = -\sum_{i=1}^N [{\rm log} p_{z}(T(x_i)) + {\rm log}\mid {\rm det} J_{T}(x)\mid_{x=x_i}].
\end{align}
Therefore, we can learn the transformation $T$ by minimizing the loss function (\ref{loss2}).

To be specific, let us briefly introduce the two transformations we will use in the following examples, i.e., neural spline flows and real-value non-volume preserving transformations (Real NVP). \\
1. Neural spline flows:\\
Durkan et al. \cite{NSF} constructed the transformation as follows,
\begin{equation}\label{NSF_trans}
\begin{split}
   \theta &= NN(x), \\
    z&=T_{\theta}(x),
\end{split}
\end{equation}
where $NN$ is a neural network and $T_{\theta}$ is a monotonic rational-quadratic spline. The parameter vector $\theta$ is used to determine the monotonic rational-quadratic spline.\\
2. Real NVP:\\
Dinh et al. \cite{RealNVP} proposed the following transformation,
\begin{equation}\label{RealNVP_Trans}
\begin{split}
z_{1}&=x_{1},  \\
z_{2}&=\frac{1}{C}[x_{2} e^{\mu(x_{1};\theta)}+\nu(x_{1};\theta)],
\end{split}
\end{equation}
where the notation $\mu$ and $\nu$ are two different neural networks. Here $C$ is a hyperparameter. We denote this transformation by $T_\theta$. 

For these two transformations, they are invertible and differentiable. The given sample data can be used to train the neural networks in the transformation. Therefore, we can estimate the target density $p_{x}(x)$ by the transformation $T_{\theta}$ and the prior density $p_{z}(z)$
\begin{align}
    p_{x}(x)&=p_{z}(T(x))\mid {\rm det} J_{T_{\theta}}(x)\mid.
\end{align}
For more details, see \cite{NSF,RealNVP,luy2021a,LuYB2021}.

Once we have the estimated density, we can use it to calculate the drift term of stochastic differential equation by the formula (\ref{T3}).

\section{Examples}
\label{exam}

Let $t^*$ be a value of time "near" zero, a dataset $\mathcal{D}=\{x_{i}^{(z)}\}_{i=1}^{n}$ sampled from $p_{x}(x,t^*|z,0)$. To be specific, these samples come from simulating stochastic differential equation with initial value $z$ using Euler-Maruyama scheme, where $z\in[-2.5,2.5]$ in one-dimensional system or $z\in[-2,2]\times[-2,2]$ in two-dimensional system.

\subsection{A one-dimensional system}
Consider a one-dimensional stochastic differential equation with pure jump L\'evy motion
\begin{align}\label{EX1}
&dX(t)=(4X(t)-X^3(t))dt+dL^{\alpha}(t),
\end{align}
where $L^{\alpha}$ is a scalar, real-value $\alpha$-stable L\'evy motion with triple $(0,0,{\nu}_{\alpha})$. The stability parameter is chosen as ${\alpha}=1.5$. The jump measure is given by
\begin{align}
\nu_{\alpha}(dx)=c(1,\alpha)|y|^{-1-\alpha}dx,
\end{align}
with $c(1,\alpha)=\frac{\alpha\Gamma((1+\alpha)/2)}{2^{1-\alpha}\pi^{\frac{1}{2}}\Gamma(1-\alpha/2)}$ as in subsection \ref{Levy}. Here we take $t^*=0.001$, sample size $n=5000$ and standard normal density as our prior density. For the transformation of normalizing flows, we take neural spline flows \cite{NSF} as our basis transformation, denoted by $T_{\theta}$. In order to improve the complexity of transformation, we take the $N=5$ compositions $T_{\theta}^N$ as our final transformation, where the neural network is a full connected neural network with 3 hidden layers and 32 nodes each layer. In addition, we choose the interval $[-B, B]=[-3,3]$ and $K=5$ bins, see \cite{NSF,luy2021a} for more details.

\begin{figure}
\centering

\includegraphics[trim={0cm 0cm 0cm 0cm},clip,width=\textwidth]{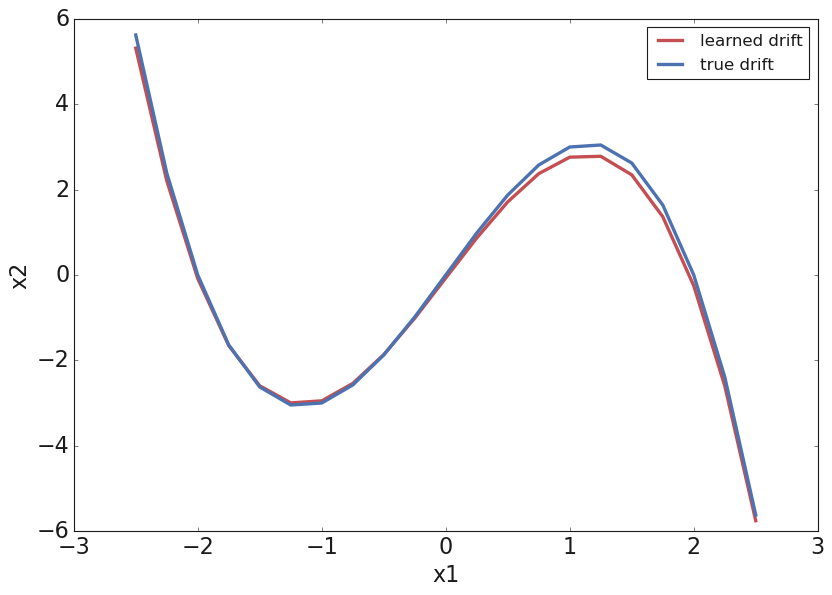}
\caption{1D system: Learning the Drift coefficient. The blue line is the true drift coefficient, and the red line is the learned drift coefficient.}
\label{1D_drift}
\end{figure}

We compare the true drift coefficient and the learned drift coefficient in Figure \ref{1D_drift}. It is seen that they are consistent with each other perfectly. In addition, the stability parameter and noise intensity are estimated as $\alpha=1.51, \sigma=0.99$ via equations (\ref{alpha}) and (\ref{sigma}), where the true values $\alpha=1.5, \sigma=1$.

\subsection{A coupled system}\label{example2}

Consider a two-dimensional stochastic differential equation with a pure jump L\'evy motion
\begin{equation}\label{EX2}
\begin{split}
&dX_{1}(t)=(5X_{1}(t)-X_{2}^2(t))dt+dL_{1}^{\alpha}(t),  \\
&dX_{2}(t)=(5X_{1}(t)+X_{2}(t))dt+dL_{2}^{\alpha}(t),\\
\end{split}
\end{equation}
where $(L_{1}^{\alpha},L_{2}^{\alpha})$ is a two-dimensional rotational symmetry real-value $\alpha-$stable L\'evy motion with triple $(0,0,\nu_{\alpha})$. The stability parameter $\alpha=1.5$. The jump measure is given by
\begin{align}
\nu_{\alpha}(dx)=c(2,\alpha)|y|^{-2-\alpha}dx,
\end{align}
with $c(2,\alpha)=\frac{\alpha\Gamma((2+\alpha)/2)}{2^{1-\alpha}\pi^{\frac{1}{2}}\Gamma(1-\alpha/2)}$. Here we take $t^*=0.001$, sample size $n=5000$ and standard normal density as our prior density. For the transformation of normalizing flows, we take RealNVP \cite{RealNVP} as our basis transformation, denoted by $T_{\theta}$. In order to improve the complexity of transformation, we take the $N=6$ compositions $T_{\theta}^N$ as our final transformation, where the neural network is a full connected neural network with 3 hidden layers and 16 nodes each layer. It should be noted that, for the RealNVP transformation (\ref{RealNVP_Trans}), the coordinate $x_1$ is identical. Therefore, we have to flip the coordinates to improve the expression of the final transformation when we compound these six basis transformations,  see \cite{RealNVP,luy2021a} for more details.

\begin{figure}
\centering
\includegraphics[trim={0cm 0cm 0cm 0cm},clip,width=\textwidth]{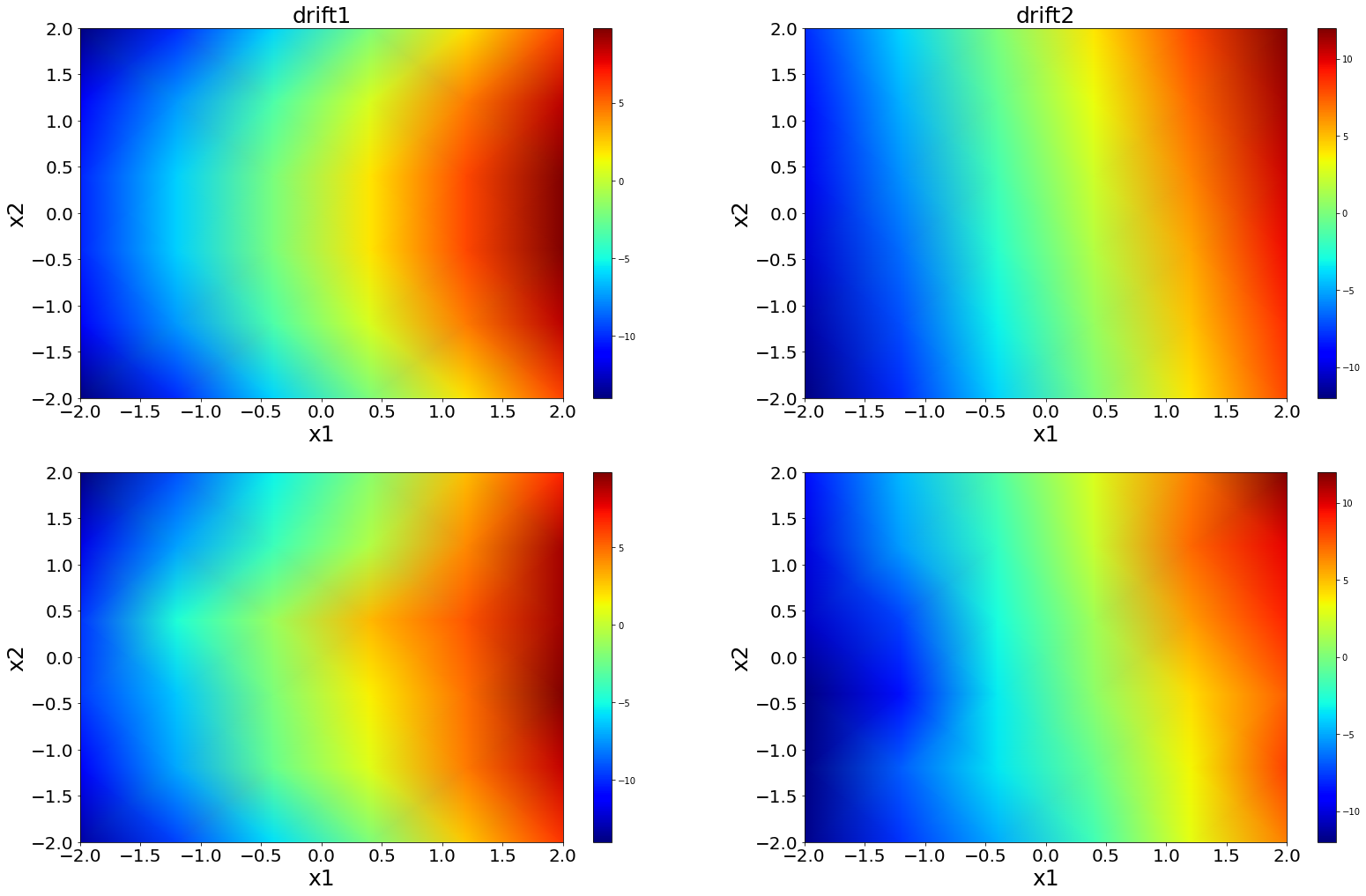}
\caption{2D coupled system with pure jump L\'evy motion: Learning the drift coefficients. Top row: The true values of drift coefficients. Bottom row: The learned values of drift coefficients.}
\label{2D_drift_coupled}
\end{figure}

We compare the true drift coefficients and the learned drift coefficients in Figure \ref{2D_drift_coupled}.  In addition, the estimated $\alpha=1.58, \sigma=0.92$, where the true values $\alpha=1.5, \sigma=1$. The learned results also agree with the true values well.

\section{Discussion}
\label{disc}
In this article, we have devised a novel data-driven method to extract stochastic dynamical systems with $\alpha$-stable L\'evy noise from short burst data. In particular, we approximate the L\'evy jump measure (stability parameter) and noise intensity through computing the mean and variance of the amplitude of increment of the sample paths. Then we estimate the drift coefficient via combining nonlocal Kramers-Moyal formulas in our previous work with normalizing flow technique. Numerical experiments on one- and two-dimensional systems illustrate the accuracy and effectiveness of our method.

Compared with our previous method, this method has the advantages that it requires less data and the algorithms are also simpler. For one-dimensional system, this method can be generalized to handle asymmetric L\'evy noise as long as we derive the third-order moment of the amplitude of rotationally symmetric random variable by its moment generating function (\ref{framo}). Then three equations can be used to calculate the stability parameter $\alpha$, the noise intensity $\sigma$ and the skewness parameter $\beta$.

Finally, note that there still exist some limitations on the applications of this method. For example, it can not be used to deal with the large jump case, i.e., $0<\alpha \le 1$. In addition, it will also cease to be effective with the existence of Brownian motion. How to solve these challenges is our future work.

\section*{Acknowledgement}
This research was supported by the National Natural Science Foundation of China (Grants 11771449).

\section*{Data Availability Statement}
The data that support the findings of this study are openly available in GitHub \url{https://github.com/liyangnuaa/Extracting-stochastic-dynamical-systems-with-alpha--stable-L-evy-noise-from-data/tree/main}.


%
%


\bibliographystyle{abbrv}
\bibliography{references}

%
%


\end{document}